\pdfoutput=1
\documentclass[11pt]{article}

\usepackage{EMNLP2022}

\usepackage{times}
\usepackage{latexsym}

\usepackage{todonotes}

\usepackage[T1]{fontenc}
\usepackage[utf8]{inputenc}

\usepackage{microtype}

\usepackage{inconsolata}

\usepackage{todonotes}
\usepackage{booktabs}
\usepackage{multirow}
\usepackage{amssymb}
\newcommand{\projname}{\textsc{Hlv}}
\usepackage{pdfcomment}

\title{The ``Problem'' of Human Label Variation: \\On Ground Truth in Data, Modeling and Evaluation}

\author{Barbara Plank\\
Center for Information and Language Processing (CIS), MaiNLP lab, LMU Munich, Germany \\
Munich Center for Machine Learning (MCML), Munich, Germany \\
\texttt{b.plank@lmu.de}}

\begin{document}
\maketitle
\begin{abstract}
Human variation in labeling is often considered noise. Annotation projects for machine learning (ML) aim at minimizing human label variation, with the assumption to maximize data quality 
and in turn optimize and maximize machine learning metrics. However, this
conventional practice assumes that there exists a \textit{ground truth}, and neglects that there exists genuine human variation in labeling due to disagreement, subjectivity in annotation or multiple plausible answers.
In this position paper, we argue that this big open problem of \textit{human label variation} persists and critically needs more attention to move our field forward. This is because human label variation impacts all stages of the ML pipeline: \textit{data, modeling and evaluation}. However, few works consider all of these dimensions jointly; and existing research is fragmented.  We reconcile different previously proposed notions of human label variation, provide a repository of publicly-available datasets with un-aggregated labels, depict approaches proposed so far, identify gaps and suggest ways forward. As datasets are becoming increasingly available, we hope that this synthesized view on the ``problem'' will lead to an open discussion on possible strategies to devise fundamentally new directions. 
\end{abstract}

\section{Introduction}

In Natural Language Processing (NLP) much progress today is driven by fine-tuning large pre-trained language models using an annotated  dataset, assumed to be representative for a target language task of interest~\cite{schlangen-2021-targeting}. This is analogously so in Machine Learning (ML) and Computer Vision (CV), where the target tasks  differ, yet the conceptual pipeline remains the same: data, modeling, evaluation. Despite the \textit{importance of annotated data}---as it fuels all steps in this pipeline---a crucial assumption of today's learning systems is to rely on a single gold label per instance. The gold label is obtained by aggregation (e.g.\ majority vote) 
of labels crucially provided by \textit{humans}. 

%while the current leading paradigm neglects genuine human disagreement, which may also reflect uncertainty in labeling, and instead continues to rely on a si C
\begin{figure}
    \centering
    \pdftooltip{\includegraphics[width=0.99\columnwidth]{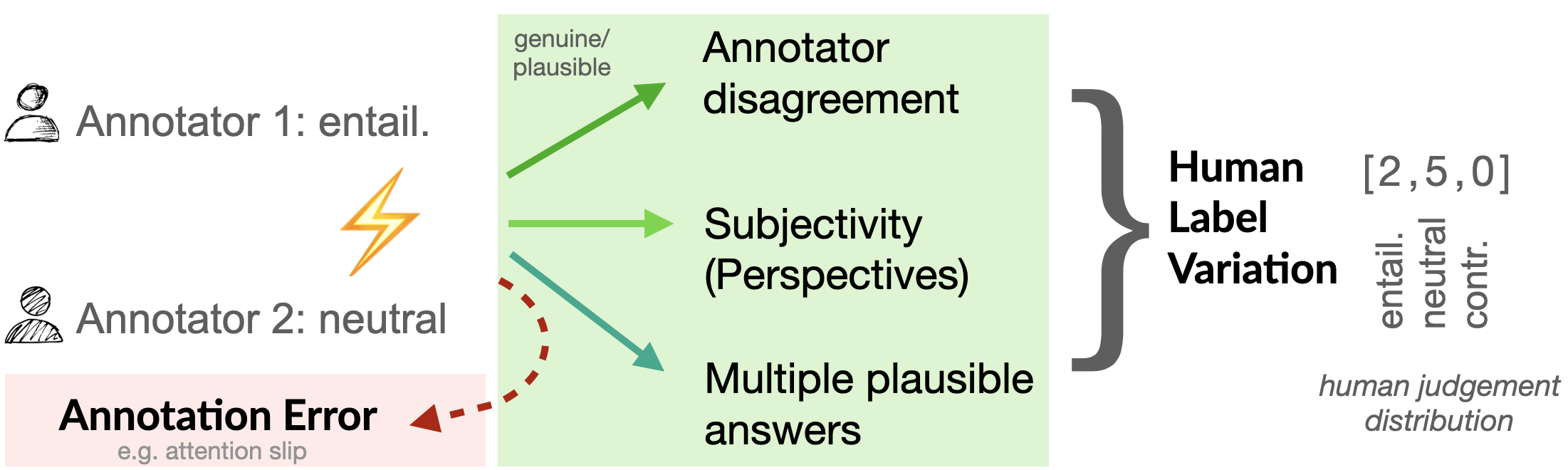}
    }{Screenreader caption: Schematic of human label variation vs annotation errors. Disagreement in annotation can be due to genuine annotator disagreement, subjectivity or simply because two (or more) views are plausible. }
    \caption{We propose the term \textit{human label variation} to capture the fact that inherent disagreement in annotation can be due to genuine disagreement, subjectivity or simply because two (or more) views are plausible.}
    \label{fig:hlv}
\end{figure}

The assumption of a \textit{ground truth} (and taking the majority vote or the `mode' of the human judgement distribution) makes sense when humans involved in labeling highly agree on the answer to the questions, such as ``Does this image contain a bird?'', ``Is `learn' a verb?'', ``What is the capital of Italy?". However, this assumption often does not make sense---especially when language is involved. For example, on questions determining a word sense, 
questions such as ``Is this comment toxic?'' or  questions involving understanding indirect answers to polar questions like ``Q:  Hey. Everything ok?'' ``A: I’m just mad at my agent'' (see more examples in Figure~\ref{fig:hardcases}). %parts of speech as in ``Is `social' a noun or adjective in `Social media are overwhelming'?''.\todo{keep POS example?} 
While some disagreement is due to human labeling errors (cf. Figure~\ref{fig:hlv} arrow to the left and \S~\ref{sec:modeling}), an increasing body of work has shown that irreconcilable variation between annotations is plausible and abundant~\cite{plank2014linguistically,aroyo2015truth,pavlick-kwiatkowski:TACL19,uma2021learning} (illustrated in Figure~\ref{fig:hlv}). The observed variation can indeed be disagreement due to difficult cases, subjectivity or cases where multiple answers are plausible (cf.\ \S~\ref{sec:data}).
We argue that human label variation (\projname) provides rich information that should not be discarded.
Critically, to rely on a  ground truth means we tacitly agree to continue: i) to create datasets that encode a single ground truth, ii) to develop models that are optimized towards a single preferred output, and iii) to evaluate models against a single ground truth. By continuing to do so, we might ask ourselves if we are climbing the right hill--or whether continuing to model a single ground truth hampers progress.

In this position paper, we argue that neglecting variation in labeling is problematic, as it impacts \textit{all steps of the pipeline}. Traditionally, this variation has been considered a problem. We underline emerging works that instead believe this issue to be an opportunity. In fact, we believe it is essential to take human label variation into account for progress. Human labels are bound to be scarce  % \todo{does this point to the fact that human annotations are needed to ground or make sense of language?}
yet at the same time critical as they provide human interpretations \textit{and values.} Therefore, embracing it is necessary for human-facing NLP, i.e., technology which is by and for humans; inclusive and reliable.  %Therefore,  embracing human label variation, and as we discuss here---as integral part of all steps in the pipeline---is necessary. 
However, the research landscape is fragmented, and approaches often focus on either steps of the pipeline. Therefore, in this paper we focus on the three core aspects of the pipeline: data, modeling and evaluation. In particular, i) we distill some of the on-going discussions in disparate (sub-)fields and propose a unified term; ii) we present and work out suggestions for each for the future; and iii) we provide a comprehensive repository of publicly-available data sets that allow studying human label variation, and invite the community to contribute.

\section{Data and Human Label Variation} 
\label{sec:data}

High-quality data is essential for any empirical scientific inquiry and has to satisfy the requirements of validity and reliability~\cite{krippendorff2018content,pustejovsky2012natural,schlangen-2021-targeting}. %\textit{Validity} entails that the sample must be relevant for the target task~\citep{pustejovsky2012natural,schlangen-2021-targeting}. \textit{Reliability}  means that the labels are reliably correct: accurate, stable and replicable~\citep{krippendorff2018content}.
However, for almost all tasks in NLP and CV irreconcilable disagreement between annotators has been observed~\cite{uma2021learning}. In light of this, the original definition of data reliability is questionable---it assumes labels follow a given standard. We might ask which standard?

\begin{figure}
    \centering
    \pdftooltip{\includegraphics[width=0.9\columnwidth]{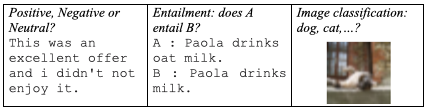}}{Screenreader caption: The figure illustrates three hard cases. Case 1: What is the sentiment (positive, neutral, negative) of `This was an excellent offer and i didn't not enjoy it'. Case 2: Does A entail B? `A: Paola drinks oat milk' `B: Paolo drinks milk.'. Case 3: It shows a low-resolution image, the question is whether it is a cat or a dog or another animal.}
    \caption{Hard cases. Image from~\cite{uma2021learning}.}
    \label{fig:hardcases}
\end{figure}

Human annotations are needed to ground and make sense of language, images, speech etc. However, labelling data is difficult, particularly when dealing with an object of study as complex as language. Take the illustration in Figure~\ref{fig:hardcases} as example. 
While categories exist, their boundaries are fluid, or simply multiple options are plausible.

\paragraph{Disagreement or variation?} We define \textit{human label variation} (\projname) as plausible variation in annotation, see Figure~\ref{fig:hlv}, to reconcile different notions found in the literature (discussed next). We prefer `variation',  because \textit{`disagreement'} implies that  two (or more) views involved cannot all hold. In contrast, errors are annotation differences,  due to amongst others attention slips.  Crucially, \projname{} assumes humans usually provide their best judgements, and variation emerges due to, e.g., ambiguity of the instance,  uncertainty of the annotator,  genuine disagreement, or simply the fact that multiple options are correct. Aggregation obfuscates this real-world complexity.

\projname{} has been studied in CV, where it is dubbed \textit{human uncertainty}~\cite{Peterson2019HumanUM}, as well as in human-computer interaction (HCI) as \textit{disagreement} or \textit{contested labels}~\cite{gordon2021disagreement}. In NLP, variation has been acknowledged as \textit{annotator disagreement} already in early works on resolving disagreement~\cite{poesio-artstein:ACL-ANNO-05}, particularly in pragmatics and discourse~\cite{de-marneffe-etal-2012-happen,webber-joshi-2012-discourse,das-etal-2017-good}. \projname{} in NLP is discussed both from the \textbf{linguistic side} as \textit{hard cases}~\cite{zeman2010hard,plank2014linguistically},~\textit{difficult linguistic cases}~\cite{manning2011part}, as judgements which are \textit{not always categorical}~\cite{de-marneffe-etal-2012-happen},
\textit{inherent disagreement}~\cite{pavlick-kwiatkowski:TACL19,davani2022dealing} and \textit{justified and informative disagreement}~\cite{sommerauer-etal-2020-describe}.
Variation in NLP is also discussed in connection to \textbf{subjectivity}, e.g., as \textit{a range of reasonable interpretations} (CrowdTruth)~\cite{aroyo2015truth}, as \textit{one or many beliefs}~\cite{paul-naacl2022}, the \textit{social dimensions of annotators} like their demographic~\cite{sap2019risk,larimore2021reconsidering,sap-etal-2022-annotators} and cultural backgrounds~\cite{hershcovich-etal-2022-challenges}, often discussed more generally as different perceptions in \textit{data perspectivisim}~\cite{basile2021toward,wich-etal-2021-investigating}. 
Moreover, there is work that acknowledges that \textbf{multiple plausible answers} are correct, such as works on the \textit{collective human opinion}~\cite{ynie2020chaosnli} influenced by seminal work that looks at the \textit{human judgement distribution}~\cite{pavlick-kwiatkowski:TACL19} who found plausible variation in at least 20\% of their data. Earlier work on veridicality also made this point~\cite{de-marneffe-etal-2012-happen}.
The fact that multiple plausible annotations exist has also been put forward as \textit{a range of acceptable annotations}~\cite{palomaki2018case}. The known variation in annotation for subjective tasks is at least a decade old~\cite{alm2011subjective}. They suggest that in the absence of a real `ground truth', acceptability may be a more useful concept than `right' and `wrong'.  Capturing the \projname{}, instead of the global majority, aligns with this viewpoint.

\paragraph{Open issues and our suggestions} To make progress, we need to i) collect and release annotator-level (un-aggregated) labels, ii) document dataset creation, and iii) include as much meta-data as possible. In particular, we urge the community to release annotator-level (un-aggregated) labels--even if only for a small subset of the data--and thus we echo~\newcite{basile2021we} and~\citet{prabhakaran-etal-2021-releasing} (also in~\citet{denton2021whose}) who independently raised this point as well. 

As a concrete starting point, we provide a comprehensive overview of existing datasets with multiple annotations in the appendix, which we release as a github repository to encourage uptake. 
Moreover, if possible to release responsibly, besides making data statements of datasets available~\cite{bender2018data}, we encourage the community to include annotator-level background information~\cite{prabhakaran-etal-2021-releasing} and document the annotation process~\cite{geiger2020garbage}. In general, we believe there is high value in releasing any meta-data available (ideally on the instance level, e.g.\ source, time of document, annotator ids, annotation completion time etc). For example, in a recent study we created a new relation extraction corpus with instance-level flags of annotator uncertainty proving valuable for evaluation~\cite{elisaspaper}. Similarly, we asked the annotator to provide free-text rationales of relations, which recently was also put forward in~\newcite{borin2022}, referring to earlier work on collecting annotator rationals during annotation~\cite{mcdonnell2016relevant}. 

We believe that the more, richer datasets become available, the more insights can be generated into the capabilities of models and their limitations. New algorithms may emerge capable of learning from fewer but richer sources. On a related line, collecting multiple annotations calls for research in estimating data quality and revisiting agreement measures; e.g., new measures for multiple-labels were recently proposed~\cite{marchal-etal-2022-establishing}.

\section{Modeling and Human Label Variation}\label{sec:modeling}

There is a growing literature on methods on how to deal with \projname{} in learning. We categorize them into two camps: those that resolve variation, and those that embrace it. We will draw connections to surveys and the emerging literature, and discuss adoption of methods as well as gaps.

The first big camp of research aims at \textbf{resolving human label variation} and includes: 1) Aggregation and 2) Filtering. It considers \projname{} as ``problematic'' or ``noisy''. Consequently, a single (aggregated) label is obtained with presumably high agreement as the ground truth. \textit{Aggregation} is performed via majority voting or probabilistic aggregation methods, see~\citet{paun2022statistical} for a survey and seminal works~\cite{dawid1979maximum,qing2014empirical,artstein2008inter}. 
Aggregation is still the most widely-adopted solution for the problem today. However, aggregation by definition allows only \textit{one belief/label/category}. This is very limiting, as often it is not just about disagreement or matter of subjectivity, but multiple options being plausible. 
 \textit{Filtering} methods are advocated by some with the idea to remove  data instances with low agreement~\cite{reidsma&carletta:CL08,reidsma&op-den-akker:08,beigman-klebanov-et-al:COLING08,beigman-beigman-klebanov-2009-learning}.  However, only using high-agreement instances can yield worse performance~\cite{jamison2015} and it wastes data.

The second camp of research instead aims at \textbf{embracing human label variation}. Two broad directions include: 3) Learning from un-aggregated labels (directly), or 4) Enriching gold with human label variation. With regard to \textit{learning from un-aggregated labels}, methods of varying complexity exist, from model-agnostic methods such as repeated labeling~\cite{sheng2008} used by e.g.~\newcite{de-marneffe-etal-2012-happen}, to architecture-specific choices, e.g., adding a crowd layer~\cite{rodrigues2018deep}, learning from soft labels~\cite{Peterson2019HumanUM} and more; see the survey of~\citet{uma2021learning}. So far learning from un-aggregated labels directly has shown greater promise in classification tasks in CV than in NLP~\cite{uma2021learning} (evidence is scarce, see open issues).  Within NLP, a more studied direction is currently \textit{to enrich the gold label with human label variation}, i.e., to learn from both the gold \textit{and} the un-aggregated labels.  Methods in this category can be seen as part of the broader set of well-known regularization methods in ML, and for NLP include e.g., cost-sensitive loss weighting~\cite{plank-et-al:EACL14:learning}, variants of multi-task learning~\cite{cohn-specia-2013-modelling,fornaciari-etal-2021-beyond,davani2022dealing}, or sequential fine-tuning~\cite{lalor2017soft}. These methods further differ in how they use un-aggregated labels, i.e.,  as confusion matrices estimated from a small sample~\cite{plank-et-al:EACL14:learning}, as annotator-level auxiliary tasks requiring the full data with multiple labels~\cite{cohn-specia-2013-modelling,davani2022dealing}, or as single ``soft-label'' auxiliary task that captures the per-instance human label distribution~\cite{fornaciari-etal-2021-beyond}. 

\begin{figure}
    \centering
    \pdftooltip{\includegraphics[width=\columnwidth]{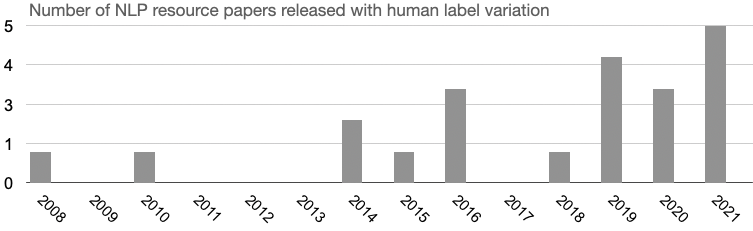}}{Screenreader caption: The graph shows a bar chat with counts of NLP paper per year that release unaggregated labels. The plot suggests an upward trend.}
    \caption{NLP Resource papers per publication year, counting publicly-available datasets released with human label variation (multiple annotator-labels per instance), cf.\ details in Table~\ref{tbl:datasets} in the Appendix.}
    \label{fig:papers}
\end{figure}

\paragraph{Open issues and our suggestions} Undoubtedly, there is increasing interest in studying methods to learn with human label variation (see Figure~\ref{fig:papers} for our analysis of research papers). 
However, existing research is fragmented across (sub)-disciplines. We identify at least three diverse areas within NLP, with little to no overlap (as shown in Table~\ref{tbl:datasets} in the Appendix), focusing respectively on: subjectivity~\cite{basile2021toward} (\url{pdai.info}, SemEval 23), natural language inference (NLI)~\cite{pavlick-kwiatkowski:TACL19,ynie2020chaosnli}, and both NLP and CV (JAIR \& SemEval 21). To the best of our knowledge, only the latter work and shared task so far bridges across disciplines~\cite{uma2021learning,uma-etal-2021-semeval}. Still, they focus on complementary NLP tasks to the two previous initiatives. It is thus an open issue to see whether tasks might need to have specific properties to be suitable for  one kind of method over another. A comprehensive evaluation is lacking. Studying transferability of methods across problems is another interesting open issue.

Learning from \projname{} heavily depends on data labeled with multiple annotators. In some settings, it might be difficult to obtain sizeable amounts of such data (however, as seen in Section~\ref{sec:data}, more datasets are emerging). Regarding learning, \citet{lalor2017soft} find that even small amounts of data can be helpful in a sequential fine-tuning setup, as also early work indicates~\cite{plank-et-al:EACL14:learning}. An open challenge is to find the right balance between the amount of data collected and the number of annotators. Overall, we hypothesize that the richness of information captured by human label variation has the potential to reduce data size requirements (possibly fewer instances but with more information captured in the human label distribution). It remains an open issue to connect with emerging works on learning with different amounts of annotation~\cite{zhang2021learning}, which can also lead to novel architectures. 

A related important challenge is to tease apart errors from signal~\cite[e.g.][]{reidsma&carletta:CL08,gordon2021disagreement}. Work on annotation error detection exists, cf.\ the very recent survey by~\citet{klie2022annotation} or~\citet{zhang-de-marneffe-2021-identifying}. It is though largely overlooked. This calls further for theoretical work on the notion of an what constitutes an error versus a hard case~\cite{manning2011part,webber-joshi-2012-discourse,plank2014linguistically}. This bears connections to emerging work in HCI, in particular social computing~\cite{gordon2021disagreement,gordon2022jury}, who look at the perception of system errors by humans, see also Section~\ref{sec:evaluation}, and earlier work in HCI on crowdsourcing that allows for some errors~\cite{krishna2016embracing}.

While embracing human label variation helps to regularize learning, the connection to a broader range of ML methods such as noise labeling or calibration remains highly relevant and  a source of further inspiration~\cite{goldberger2016,han2018co,han2018masking,meister2020generalized}. %Noise labeling~\cite{goldberger2016,han2018co,han2018masking} treats disagreement as a corruption of a theoretical gold standard, while calibration methods are a widely-adopted method to counter overconfidence of neural classifiers~\cite{meister2020generalized}. 
There are some initial studies that compare human disagreement with model confidence~\cite{davani2022dealing}. Overall, interest in calibration methods~\cite{naein-obtaining-2015, guo-on-calibration-2017} is increasing~\cite{desai-durrett-2020-calibration, kong-etal-2020-calibrated, jiang-how-can-2021} to counter overconfidence of neural classifiers~\cite{meister2020generalized}.  In contemporary work to this, we show that measuring calibration to human majority given inherent disagreements is theoretically and empirically problematic~\cite{jorispaper}. As a first step, we propose instance-level measures of calibration that better capture the human label distribution. In future, it remains to be seen how to best use human label variation to make systems more trustworthy. 

Finally, there is relevant interesting work that more deeply looks at data during learning. In NLP, recent seminal work by~\citet{swayamdipta-etal-2020-dataset} proposes \textit{data maps} to investigate the behavior of a model on individual instances during training (training dynamics). They show that training a system on \textit{ambiguous} instances identified via data maps helps to generalizes better in out-of-distribution evaluation~\cite{swayamdipta-etal-2020-dataset}. Building on top of this work,~\newcite{zhang-plank-2021-cartography-active} show that the instances at the boundary of hard and ambiguous cases derived from small data maps aids active learning. This is further evidence that human uncertainty in labeling is beneficial for learning. It remains to be seen whether training dynamics can yield novel architectures for learning from \projname.

\section{Evaluation and Human Label Variation}\label{sec:evaluation}

Evaluation is of critical importance in empirical research fields such as ML, NLP and CV. It helps to choose one system over another, and to measure progress. However, current evaluation practices typically use accuracy against a gold standard. In many tasks this common practice is severely flawed. It obfuscates the truth about the state of ML models. It leaves a large gap between in-vitro and in-vivo evaluation. HCI research has shown that metrics are not aligned with reality; audits of algorithms' performance have uncovered very poor results in practice, and that  this disconnect is indicative of a larger disconnect on how ML and HCI researchers evaluate their work~\cite{gordon2021disagreement}. We believe this is an important take-away for NLP. We too often focus on single metrics, single components of the pipeline, in other words, on myopic \textit{in-vitro} experimentation.

\paragraph{Open issues and our suggestions} Despite the increasing body of literature on methods for learning with \projname, a majority of the papers introducing new methods strikingly evaluate against \textit{hard labels} (gold labels)~\cite[e.g.][]{rodrigues2018deep,fornaciari-etal-2021-beyond}. If we want to take human label variation seriously, we need to shift our attention to evaluation that goes beyond hard labels (accuracy). As accuracy of all models can be high (at times), looking at only one metric (and, in fact a single---argmax---prediction) gives no indication on how reasonable a model is, yet alone how confident and trustworthy it is.

Research in ML, CV and NLP has started to incentivize  \textit{hard and soft label} evaluation. Soft labels compare the human label distribution to model outputs. Proposed soft metrics include: \textit{cross entropy}, to capture how well the model captures humans' assessment not just of the top label, which is used in both CV~\cite{Peterson2019HumanUM} and NLP~\cite{pavlick-kwiatkowski:TACL19}; \textit{entropy correlation} proposed by~\citet{uma-et-al:HCOMP20}, to compute Pearson's correlation between instance-level  entropy scores of human soft labels and model predictions;  \textit{Kullback-Leibler divergence}-based evaluation~\cite{ynie2020chaosnli} (either KL or Jensen-Shannon). Others instead started to evaluate against \textit{individual annotators}~\cite{resnick2021survey,davani2022dealing}, measure F1 scores against \textit{data splits by different annotator agreement levels}~\cite{leonardelli-etal-2021-agreeing,damgaard-etal-2021-ill}, data splits based on \textit{annotator clustering}~\cite{basile2021toward}, data splits based on \textit{item difficulty} based on entropy of the label distribution and semantic distance~\cite{jolly-etal-2021-ease}, and data splits based on \textit{annotator uncertainty flags}~\cite{elisaspaper}. Analogously as in Section~\ref{sec:modeling},  it is an open issue to see whether tasks might need to have specific properties to be more suitable for one kind of evaluation over another. In general, we need better evaluation practices (besides soft and hard evaluation), particularly in light of the complexity of human label variation---and the reasons it arises, which might be due to uncertainty, background, task complexity, intra-coder reliability etc; see~\newcite{basile2021we} and in particular~\newcite{jiang-marneffe2022} for a discussion on disagreement sources; the latter recently developed a taxonomy for disagreement in natural language inference data.

\section{Conclusions}
In this paper, we outline that human label variation impacts all steps of the traditional ML pipeline, and is an opportunity, not a problem. To move forward, we argue for a more comprehensive treatment of \projname{}, which considers all steps, to enable innovation: data, modeling and evaluation. To do so, and truly move beyond the current in-vitro setups, we need an open, inter-disciplinary discussion. We hope to contribute to this discussion, and stipulate research with the released repository: \url{https://github.com/mainlp/awesome-human-label-variation}.\footnote{The repository contains the datasets in Appendix 1 as a starting point. This is, to the best of our knowledge, the most comprehensive list of datasets with un-aggregated labels available today. We encourage readers to contribute. They are further invited to join the SemEval 2023 shared task LeWiDi: \url{https://le-wi-di.github.io/}}

%\newpage
\section*{Limitations}

This position paper tries to be succinct while aiming at synthesizing a very broad notion---human label variation---that affects all steps dealing with learning from annotated data. Therefore, this position paper is necessarily incomplete, as is the dataset repository that is provided. However, we hope that the repository and paper will lead to an open discussion and community uptake, as this is a big open issue and necessitates a broader, inter-disciplinary treatment.

\section*{Ethics Statement}
Modeling human label variation is connected to social bias, as annotator backgrounds influence annotations and consequently both machine learning and evaluation. Therefore it is important to be aware of possible social implications of some of the technologies discussed here. Inevitably there is potential for dual use, as amplifying the voice of some might harm others. However, there are  social opportunities, as modeling human label variation allows to include the voices of more groups, and even the very underrepresented. In a world where the majority view dominates, these would otherwise be left behind.

\section*{Acknowledgements}
We thank the reviewers for their feedback. Special thanks to Bonnie Webber, Massimo Poesio and Raffaella Bernardi for invaluable feedback on drafts of this paper and discussions on this topic, in parts at the Insights workshop@ACL 2022. Thanks to members of the NLPnorth \& MaiNLP lab for feedback on this paper.
BP is supported by ERC Consolidator Grant DIALECT 101043235.

\bibliography{anthology,custom}
\bibliographystyle{acl_natbib}

\appendix

\section{Datasets with Multiple Annotations}
\label{sec:appendix-data}

\begin{table*}[h!]
    \centering
    \resizebox{\textwidth}{!}{
    \begin{tabular}{l|p{4cm}p{7cm}p{5.5cm}ccccc}
    \toprule
    \textbf{Field} &  \textbf{Reference}   & \textbf{Name/Description}  & \textbf{URL} & \textbf{Jair} & \textbf{PDAI} & \textbf{SemEval21} & \textbf{TACL22} & \textbf{SemEval23} \\
     \midrule
     \multirow{19}{*}{NLP} & \cite{passonneau2010word} & Word sense disambiguation (WSD) & \small{\url{https://anc.org/}}\\
     &  \cite{plank-et-al:EACL14:learning}   &  POS tagging (500 tweets from Lowlands) and Gimpel-POS dataset & \small{\url{https://bitbucket.org/lowlands/costsensitive-data/}} and \small{\url{https://zenodo.org/record/5130737}} & $\checkmark$ & & $\checkmark$ \\% \url{https://competitions.codalab.org/competitions/25748}  \\
     & \cite{derczynski-etal-2016-broad} & NER Broad Twitter dataset & \small{\url{https://github.com/GateNLP/broad_twitter_corpus}} & & $\checkmark$ &\\
     & \cite{rodrigues2018deep} & NER dataset, re-annoted sample of CoNLL 2003 & \small{\url{http://fprodrigues.com//publications/deep-crowds/}}\\
     & \cite{martinez-alonso-etal-2016-supersense} & Supersense tagging & \small{\url{https://github.com/coastalcph/semdax}} \\
     &  \cite{berzak-etal-2016-anchoring} & Dependency Parsing, WSJ-23, 4 annotators & \small{\url{https://people.csail.mit.edu/berzak/agreement/}} \\
         &  \cite{logan-aacl} & GCDT, Mandarin Chinese discourse treebank & \small{\url{https://github.com/logan-siyao-peng/GCDT/tree/main/data}} \\
     & \cite{bryant-ng-2015-far} & Grammatical error correction & \small{\url{http://www.comp.nus.edu.sg/~nlp/sw/10gec_annotations.zip}} \\
 %   & \cite{} & Dependency Parsing & \small{\url{https://bitbucket.org/lowlands/release/src/master/CoNLL2015/datapackage/}} \\
     & \cite{poesio-etal-2019-crowdsourced} & PD (Phrase Detectives dataset): Anaphora and Information Status Classification & \small{\url{https://github.com/dali-ambiguity/Phrase-Detectives-Corpus-2.1.4}} & $\checkmark$ & & $\checkmark$\\
     & \cite{dumitrache-etal-2018-crowdsourcing} & Medical Relation Extraction (MRE) & \small{\url{https://github.com/CrowdTruth/Open-Domain-Relation-Extraction}} & $\checkmark$ &\\
     & \cite{elisaspaper} & CrossRE, relation extraction, small doubly-annotated subset & \small{\url{https://github.com/mainlp/CrossRE}} & \\
     & \cite{dumitrache2018frames} & Frame Disambiguation & \small{\url{https://github.com/CrowdTruth/FrameDisambiguation}}\\
     & \cite{snow2008cheap} & RTE (recognizing textual entailment; 800 hypothesis-premise pairs) collected by~\cite{dagan2005pascal}, re-annotated; includes further datasets on temporal ordering, WSD, word similarity and affective text  & \small{\url{https://sites.google.com/site/nlpannotations/}} & $\checkmark$ &\\
     & \cite{pavlick-kwiatkowski:TACL19} & NLI (natural language inference) inherent disagreement dataset, approx.\ 500 RTE instances from~\cite{dagan2005pascal} re-annotated by 50 annotators & \small{\url{https://github.com/epavlick/NLI-variation-data}}\\
     & \cite{ynie2020chaosnli} & ChaosNLI, large NLI dataset re-annotated by 100 annotators & \small{\url{https://github.com/easonnie/ChaosNLI}}\\
     & \cite{demszky-etal-2020-goemotions} & GoEmotions: reddit comments annotated for 27 emotion categories or neutral & \small{\url{https://github.com/google-research/google-research/tree/master/goemotions}} & & & & $\checkmark$\\
    & \cite{ferracane-etal-2021-answer} & Subjective discourse: conversation acts and intents & \small{\url{https://github.com/elisaF/subjective_discourse}}\\
    & \cite{damgaard-etal-2021-ill} & Understanding indirect answers to polar questions & \small{\url{https://github.com/friendsQIA/Friends_QIA}}\\ 
    & \cite{de2019commitmentbank} & CommitmentBank:  8 annotations indicating the extent to which
the speakers are committed to the
truth of the embedded clause &  \small{\url{https://github.com/mcdm/CommitmentBank}}\\
    & \cite{kennedy2020constructing} & Hate speech detection & \small{\url{https://huggingface.co/datasets/ucberkeley-dlab/measuring-hate-speech}} & & $\checkmark$ & & $\checkmark$\\
     & \cite{dinu-etal-2021-computational-exploration} & Pejorative words dataset & \small{\url{https://nlp.unibuc.ro/resources}} or \small{\url{http://pdai.info/}} & & $\checkmark$ &\\
     & \cite{leonardelli-etal-2021-agreeing} &  MultiDomain Agreement, Offensive language detection on Twitter, 5 offensive/non-offensive labels; also part of LeWiDi SemEval23  & \small{\url{https://github.com/dhfbk/annotators-agreement-dataset/}} & & $\checkmark$ & & & $\checkmark$ \\
     & \cite{cercas-curry-etal-2021-convabuse} & ConvAbuse, abusive language towards three conversational AI systems; also part of LeWiDi SemEval23 & \small{\url{https://github.com/amandacurry/convabuse}} & & $\checkmark$ & & & $\checkmark$\\
     & \cite{Liu2019ACL} &  Work and Well-being Job-related Tweets, 5 annotators & \small{\url{https://github.com/Homan-Lab/pldl_data}}& & $\checkmark$ &\\
     & \cite{simpson-etal-2019-predicting} & Humour: pairwise funniness judgements & \small{\url{https://zenodo.org/record/5130737}} &  & & $\checkmark$\\
     & \cite{akhtar2021whose} & HS-brexit; New LeWiDi-23 shared tast dataset on Abusive Language on Brexit and annotated for hate speech (HS), aggressiveness and offensiveness, 6 annotators & \url{https://le-wi-di.github.io/} & & $\checkmark$ & & & $\checkmark$\\
     & \cite{almanea-poesio:ARMIS} & ArMIS; New LeWiDi-23 shared tast dataset on Arabic tweets annotated for misogyny detection & \url{https://le-wi-di.github.io/} & & &  & & $\checkmark$\\
     
     \midrule
    
     \multirow{2}{*}{CV} & \cite{rodrigues2018deep} & LabelMe: Image classification dataset with 8 categories, re-annotated   & \small{\url{http://fprodrigues.com//publications/deep-crowds/}} & $\checkmark$ & & $\checkmark$\\
     & \cite{Peterson2019HumanUM} & Cifar10H: Image classification with 10 categories, re-annotated & \small{\url{//github.com/jcpeterson/cifar-10h}} & $\checkmark$ & & $\checkmark$ \\
    % & \cite{guan2018said} & Diabetic Retinopathy Classification & n/a \\
     & \cite{cheplygina2018crowd} & Medical lesion classification challenge, 6 annotators each & \small{\url{https://figshare.com/s/5cbbce14647b66286544}}\\
     \bottomrule
    \end{tabular}
    }
   
    \caption{Overview of publicly-available datasets with \projname{} data (see repository for updates and to contribute: \url{https://github.com/mainlp/awesome-human-label-variation}). $\checkmark$: whether the source was used in broader empirical evaluations, e.g., the JAIR survey on learning from disagreement~\cite{uma2021learning}, is listed on \url{pdai.info}~\cite{basile2021toward} (as of June, 2022), is part of the SemEval 2021 task on learning from disagreement~\cite{uma-etal-2021-semeval}, is used in a TACL paper on learning beyond majority vote~\cite{davani2022dealing}, is used in the SemEval 2023 shared task on Learning With Disagreement LeWiDi \url{https://le-wi-di.github.io/}.}
    
    \label{tbl:datasets}
\end{table*}

%This is an appendix.

\end{document}